\documentclass[runningheads]{llncs}

\usepackage{graphicx}
\usepackage{enumitem}

\usepackage{multirow}
\usepackage{color}

\usepackage{makeidx, tabularx, graphicx, setspace, amsmath, longtable, multirow, float, theorem, times, booktabs, siunitx, lineno}

\usepackage{verbatim} 

\begin{document}

\title{When Unseen Domain Generalization is Unnecessary? Rethinking Data Augmentation}

\author{Ling Zhang\inst{1} \and Xiaosong Wang\inst{1} \and Dong Yang\inst{1} \and Thomas Sanford\inst{3} \and Stephanie Harmon\inst{4} \and Baris Turkbey\inst{3} \and Holger Roth\inst{1} \and Andriy Myronenko\inst{2} \and Daguang Xu\inst{1} \and Ziyue Xu\inst{1}}

\authorrunning{L. Zhang, et al.}
\titlerunning{Unseen Domain Generalization}

\institute{NVIDIA, Bethesda MD 20814, USA \and NVIDIA, Santa Clara CA 95051, USA \and NIH, Bethesda MD 20892, USA \and NCI, Bethesda MD 20892, USA}

%
\maketitle              
\begin{abstract}
Recent advances in deep learning for medical image segmentation demonstrate expert-level accuracy. However, in clinically realistic environments, such methods have marginal performance due to differences in image domains, including different imaging protocols, device vendors and patient populations.
Here we consider the problem of domain generalization, when a model is trained once, and its performance generalizes to unseen domains.  Intuitively, within a specific medical imaging modality the domain differences are smaller relative to natural images domain variability. We rethink data augmentation for medical 3D images and propose a deep stacked transformations (DST) approach for domain generalization. Specifically, a series of $n$ stacked transformations are applied to each image in each mini-batch during network training to account for the contribution of domain-specific shifts in medical images. 
We comprehensively evaluate our method on three tasks: segmentation of whole prostate from 3D MRI, left atrial from 3D MRI, and left ventricle from 3D ultrasound.
We demonstrate that when trained on a small source dataset, (i) on average, DST models on unseen datasets degrade only by 11\% (Dice score change), compared to the conventional augmentation (degrading 39\%) and  CycleGAN-based domain adaptation method  (degrading 25\%), (ii) when evaluation on the same domain, DST is also better albeit only marginally. 
(iii) When training on large-sized data, DST on unseen domains reaches performance of state-of-the-art fully supervised models.
These findings establish a strong benchmark for the study of domain generalization in medical imaging, and can be generalized to the design of robust deep segmentation models for clinical deployment.

\end{abstract}

\section{Introduction}
\begin{figure*}[!t]
   \vspace{-5pt}
   \begin{center}
   \begin{tabular}{c}
   \includegraphics[width=12cm]{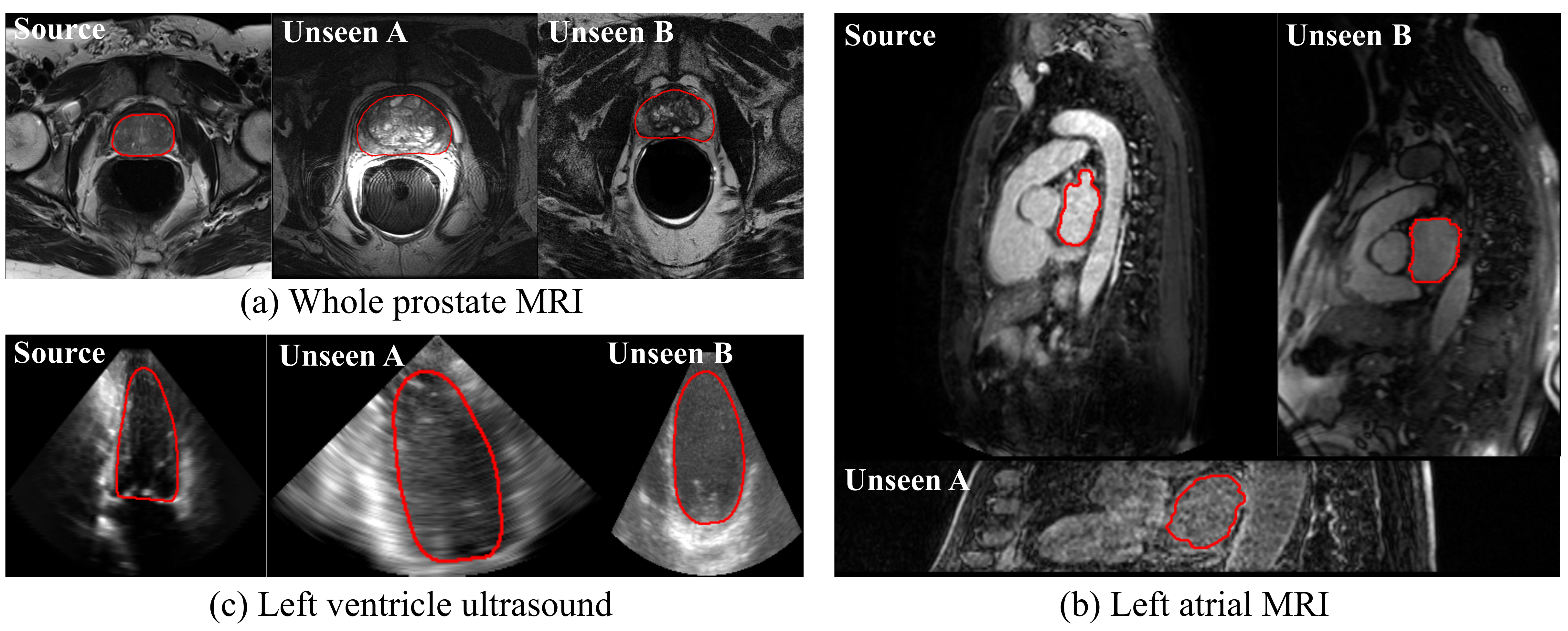}
   \end{tabular}
   \vspace{-10pt}
   \end{center}
   \setlength{\belowcaptionskip}{-5pt}
   \caption
   { \label{fig1} 
Medical image segmentation in source and unseen domains (i.e., a specific medical imaging modality across different vendors, imaging protocols, and patient populations, etc.) for (a) whole prostate MRI, (b) left atrial MRI, and (c) left ventricle ultrasound. The illustrated images are processed with intensity normalization.
}  
   \end{figure*}

Practical application of AI medical imaging methods require accurate and robust performance on unseen domains, such as differences in acquisition protocols across different centers, scanner vendors, and patient populations (see Fig.~\ref{fig1}). Unfortunately, labeled medical datasets are typically small and do not include sufficient variability for robust deep learning training.  
The lack of large, diverse medical imaging datasets often lead to marginal deep learning model performance on new ``unseen" domains, which limits their applications in clinical practice~\cite{yasaka2018deep}.

To improve model performance on unseen domains, transfer learning methods attempt to fine-tune a portion of a pre-trained network given a small amount of annotated data from the unseen target domain. Transfer learning applications for medical 3D images are often lacking quality pre-trained models (trained on a large amount of data). Domain adaption methods do not require annotations in the unseen domain, but usually require all source and target domain images be available during training~\cite{zhu2017unpaired,degel2018domain,zhang2018task}. 
The assumption of a known target dataset is restrictive, and makes multi-site deployment impractical. Furthermore, due to medical data privacy requirements, it is difficult to collect both the source and target datasets beforehand.

In the field of medical imaging, we are usually faced with the difficult situation where the training dataset is derived from a single center and acquired with a specific protocol. In such situations, domain generalization methods seek a robust model, trained once, capable of generalizing well to unseen domains. In 2D computer vision applications, researchers focused on various complexity of  data augmentation to expand the available data distribution. Specifically, data augmentation strategies are performed in input space~\cite{romera2018train} or during adversarial learning~\cite{volpi2018generalizing}.
Compared to natural 2D images, 3D medical image domain variability is more compact. Within the same modality, e.g. T2 MRI or Ultrasound, images from different vendors (GE, Philips, Siemens), scanning protocols, and patient populations are visually different mainly in three aspects: image quality, image appearance and spatial shape (see Figure~\ref{fig1}). Other imaging modalities, such as CT, generally have more consistent image characteristics.    


Motivated by the observed heterogeneity of 3D medical images, we propose a systematic augmentation approach consisting of series of transformations to simulate domain shift properties of medical imaging data. We call this approach, Deep Stacked Transformations (DST) augmentation. 
DST operates on the image space, where input images undergo nine stacked transformations. Each transform is controlled by two parameters, which determine the probability and magnitude of the image transformation. As a backbone semantic segmentation network we use AH-Net~\cite{liu20183d}.  

In 3D medical imaging applications, the selection of image augmentations is often intuitive, random crop or flip, inherited from 2D computer vision applications. Furthermore, the contribution of augmentation method is rarely evaluated on the unseen domain.
In this work, we comprehensively evaluate the effect of various data augmentation techniques on 3D segmentation generalization to the unseen domains. The evaluation tasks include segmentation of whole prostate from 3D MRI, left ventricle from 3D ultrasound, and left atrial from 3D MRI. For each task we have up to 4 different datasets to be able to train on one and evaluate generalization to other datasets. The results and analysis
\begin{itemize}[label=$\bullet$]
\item Reveal the main factors causing domain shift in 3D medical imaging modalities.


\item Demonstrate that DST augmentation substantially outperforms conventional augmentation and CycleGAN-based domain adaptation on unseen domains for both MRI and ultrasound. 
The generalization improvements are observed even on the same domain (albeit much less noticeable).

\item 
Given a larger training dataset, DST achieves state-of-the-art segmentation accuracy on unseen domains.



\end{itemize}

\section{Methods}

\begin{figure*}[!t]
   \vspace{-5pt}
   \begin{center}
   \begin{tabular}{c}
   \includegraphics[width=11.8cm]{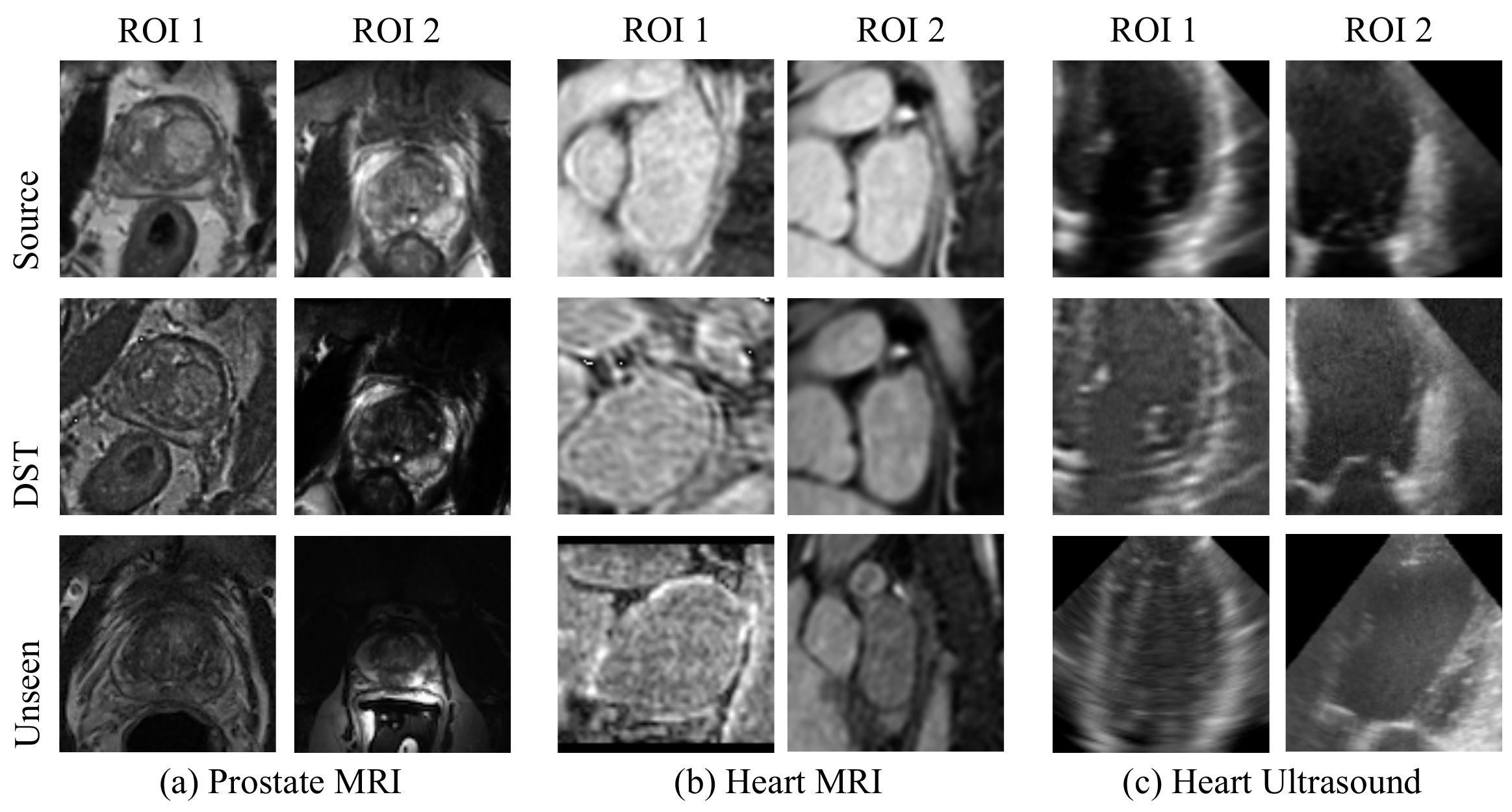}
   \end{tabular}
   \vspace{-12pt}
   \end{center}
   \caption
   { \label{fig_aug} 
Examples of deep stacked transformations (DST) results on (a) whole prostate MRI, (b) left atrial MRI, and (c) left ventricle ultrasound. 1st row: ROIs randomly cropped from source domains; 2nd row: corresponding ROIs after DST; 3rd row: ROIs randomly cropped from unseen domains. The image pairs of 2nd--3rd rows have better visual similarity than 1st--3rd rows.
}  
   \end{figure*}

To improve generalization of 3D medical semantic segmentation method, we use a series $n$ stacked augmentation transforms $\tau(.)$ applied to input images during training.  Each transformation is an image processing function with two hyper-parameters: probability $p$ and magnitude $m$.
\begin{equation}
	(\hat{x}_{s}, \hat{y}_{s}) = \tau_{p_{n}, m_{n}}^{n} ( \tau_{p_{n-1}, m_{n-1}}^{n-1} (... \tau_{p_{1}, m_{1}}^{1} (x_{s}, y_{s}) ) )
	\label{9transforms}
\end{equation}
where  $x_{s}, y_{s}$ are input image and its corresponding label.
Augmentation transforms alter the image quality, appearance, and spatial structure. Specifically DST consists of the following transforms: sharpening, blurring, noise, brightness adjustment, contrast change, perturbation, rotation, scaling, deformation, in addition to random cropping. In DST, transforms are in the order as described -- performances of models are not sensitive to different orders.
As we show in our experiments, augmenting image sets during training can result in models with more robust segmentations than if data processing/synthesis was performed at the inference stage. 
Fig.~\ref{fig_aug} shows some examples of DST augmentation in 3D MRI and ultrasound demonstrating ability to mimic image appearances in unseen domains with a given modality. 

\noindent\textbf{Image Quality} is related to \textit{sharpness}, \textit{blurriness}, and \textit{noise level} of medical images. Blurriness is commonly caused by MR/ultrasound motion artifacts and resolution. Gaussian filtering is used to blur the image, with a magnitude (Gaussian std) ranging between [0.25, 1.5]. Sharpness has a reverse effect, by using an unsharp masking with strength [10, 30]. Noise is added (from normal distribution with std. [0.1, 1.0]) to account for possible noise in images. 


\noindent\textbf{Image Appearance} is associated with the statistical characteristics of image intensities, such as variations of \textit{brightness} and \textit{contrast}, which often result form different scanning protocols and device vendors.  Brightness augmentation refers to random shift [-0.1, 0.1] in the intensity space. Contrast augmentation refers to gamma correction with gamma (magnitude) ranging between [0.5, 4.5]. Finally, we use a random linear transform in intensity space with magnitude of scale and shift sampled from [-0.1, 0.1], which we refer to as intensity \textit{perturbation}.


\noindent\textbf{Spatial Transforms} include \textit{rotation}, \textit{scaling}  and \textit{deformation}. Rotation is usually caused by different patient orientations during scanning (we use [-20$^{\circ}$, 20$^{\circ}$] range). Scaling and Deformation are due to organ shape variability and soft tissue motion. 
Random scaling is used with magnitude [0.4, 1.6]. Deformation transform uses regular grid interpolation, after a random perturbation (Gaussian smoothed std [10, 13]). Same spatial transform are applied to both input images and the corresponding labels.
These operations are computational expensive for large 3D volumetric data. GPU-based acceleration approach could be developed, but allocating the maximal capacity of GPU memory for model training only along with data augmentation on the fly are more desirable. In addition, since the whole 3D volume does not fit into the memory of the GPU, sub-volumes cropping are usually needed to fed into network training. We develop a CPU-based, efficient, spatial transform technique based on an open-source implementation\footnote[1]{https://github.com/MIC-DKFZ/batchgenerators}, which first calculates the 3D coordinate grid of sub-volume (with size of $w \times h \times d$ voxels) to which the transformations (combining random 3D rotation, scaling, deformation, and cropping) are applied and then image interpolation is performed. We make further acceleration by only performing interpolation within the minimal cuboid containing the 3D coordinate grid, as such, the computational time is independent from the input volume size (i.e., only depend on the cropping sub-volume size), and the spatial transform augmentation can be performed on the fly during training.

\section{Experiments}
 
\subsection{Datasets}
We validate our method on three segmentation tasks: segmentation of whole prostate from 3D MRI, left atrial from 3D MRI, and left ventricle from 3D ultrasound.

\noindent\textbf{Task 1:} For the whole prostate segmentation from 3D MRI, we use the following datasets: Prostate dataset from Medical Segmentation Decathlon\footnote[1]{http://medicaldecathlon.com/index.html} (MSD-P), PROMISE12 \cite{litjens2014evaluation}, NCI-ISBI13\footnote[2]{http://doi.org/10.7937/K9/TCIA.2015.zF0vlOPv}, and ProstateX \cite{litjens2014computer}. We train on the MSD-P dataset (source domain) and evaluate on the other datasets (unseen domains). We use only single channel (T2) input and segment the whole prostate, which is lowest common denominator among the datasets.  One study in ProstateX was excluded due to prior surgical procedure. 


\noindent\textbf{Task 2:} For left atrial segmentation from 3D MRI, we use the following datasets: Heart dataset from MSD$^{1}$ (MSD-H), ASC \cite{xiong2019fully} 
and MM-WHS \cite{zhuang2016multi}. We train on the MSD-H dataset (source domain) and evaluate on the other datasets.


\noindent\textbf{Task 3:} For left ventricle segmentation from 3D ultrasounds, we use data from CETUS\footnote[3]{https://www.creatis.insa-lyon.fr/Challenge/CETUS/} (30 volumes). We manually split the dataset into 3 subsets corresponding to different ultrasound device vendors A, B, C with 10 volumes each. We used heuristics to identify vendor association, but we acknowledge that our split strategy may include wrong associations.  We train on Vendor A images, and evaluate on Vendors B and C. 

Table \ref{datasets} summarizes the datasets. In addition, a larger proprietary 3D MRI dataset of 465 volumes is used in the final experiment (see Section~\ref{scalingup}).

\begin{table}[!t]
\caption{Datasets used in our experiment.}
\vspace{-5pt}
\label{datasets}
\scriptsize
\centering
\begin{tabular}{l|c|c|l|l|c|c|l|ccl}
\hline
\multicolumn{1}{c|}{Task} & \multicolumn{4}{c|}{\textbf{1.} MRI - whole prostate}                                                                       & \multicolumn{3}{c|}{\textbf{2.} MRI - left atrial}                                          & \multicolumn{3}{c}{\textbf{3.} Ultrasound - left ventricle}                                        \\ \hline
Domain                    & Source                     & \multicolumn{3}{c|}{Unseen}                                                        & Source                     & \multicolumn{2}{c|}{Unseen}                        & \multicolumn{1}{c|}{Source}   & \multicolumn{2}{c}{Unseen}                             \\ \hline
Dataset                   & \multicolumn{1}{l|}{MSD-P} & \multicolumn{1}{l|}{PROMISE12} & NCI-ISBI13              & ProstateX               & \multicolumn{1}{l|}{MSD-H} & \multicolumn{1}{l|}{ASC} & MM-WHS                  & \multicolumn{1}{l|}{CETUS-A} & \multicolumn{1}{l|}{CETUS-B} & CETUS-C               \\ \hline
\# Data                   & 26/6                       & 50                             & \multicolumn{1}{c|}{60} & \multicolumn{1}{c|}{98} & 16/4                       & 100                      & \multicolumn{1}{c|}{20} & \multicolumn{1}{c|}{8/2}      & \multicolumn{1}{c|}{10}       & \multicolumn{1}{c}{10} \\ \hline
\end{tabular}
\vspace{-8pt}
\end{table}

\subsection{Implementation}
\label{implementation}
We implemented our approach in Tensorflow and train it on NVIDIA Tesla V100 16GB GPU.  We use AH-Net~\cite{liu20183d} as a backbone for 3D segmentation, which takes advantages of the 2D pretrained ResNet50 as an encoder, and learns the full 3D decoder. All data is re-sampled to 1x1x1mm isotropic resolution and normalized to [0,1] intensity range. We use a crop size of 96x96x32 batch 16 for Task1, crop 96x96x96 batch 16 for Tasks 2, and crop 96x96x96 batch 4 for Tasks 3. 
We use soft Dice loss and Adam optimizer with the learning rate $10^{-4}$. We use 0.5 probability of each transformation in DST.  

\subsection{Experimental Results and Analysis}
First, we evaluate generalization performance for each augmentation transform individually. As a \emph{baseline}, only random cropping with no other augmentations used. We compare results to DST with all 9 transformation stacked, and to a popular domain adaptation method, CycleGAN~\cite{zhu2017unpaired}, which maps the unseen images (on per slice basis) into source-like appearance (we split each dataset into 4:1 for CycleGAN training and validation, and train for 200 epochs). 


Table~\ref{effect} lists segmentation Dice results on the \emph{source domain} (trained on this domain, and validated on a keep-out subset) and on \emph{unseen domains} (trained on the source, but tested on other unseen datasets). The major findings are:

\begin{table}[!t]
\caption{The effect of DST and various augmentation methods on unseen domain generalization (measured as segmentation Dice scores). \emph{Source} columns indicates the dataset used for training, and its Dice scores are validation Dice scores (using a split) for comparisons. \emph{Unseen} columns list Dice results when applied to unseen datasets (of the model trained on the source). Here baseline refers to a random crop with no further augmentations. \emph{Top4} stands for the combination of four best performing augmentations (sharpening, brightness, contrast, scaling). \emph{Supervised} indicates the state-of-the-art literature results, when a model is trained and tested on the same dataset. $*$ indicates inter-observer variability.} 

\label{effect}
\scriptsize
\centering

\begin{tabular}{|l|c|c|c|c|c|c|c|c|c|c|c|c|}
\hline
                 & \multicolumn{4}{c|}{Task 1. MRI - whole prostate}             & \multicolumn{3}{c|}{Task 2. MRI- left atrial} & \multicolumn{3}{c|}{Task 3. US - left ventricle} & \multicolumn{2}{c|}{All Tasks} \\ \hline
                 & Source        & \multicolumn{3}{c|}{Unseen}                   & Source        & \multicolumn{2}{c|}{Unseen}   & Source         & \multicolumn{2}{c|}{Unseen}     & Source         & Unseen        \\ \hline
                 & MSD-P         & PROMISE       & NCI-ISBI      & ProstateX     & MSD-H         & ~ASC~           & WHS           & CETUS-A        & ~~C-B~~            & C-C            & Average        & Average       \\ \hline
Baseline         & 89.6          & 60.4          & 58.0          & 76.8          & \textbf{91.9} & 4.4           & 72.9          & 85.8           & 51.7           & 39.2           & 89.1           & 49.8          \\ \hline
Sharpening       & 90.6          & 65.5          & 82.8          & 84.0          & 91.5          & 5.7           & 78.9          & 83.7           & 59.5           & 78.5           & 88.6           & 62.9          \\ \hline
Blurring         & 86.1          & 63.9          & 67.0          & 79.9          & 90.9          & 3.3           & 76.9          & 90.5           & 73.4           & 72.4           & 89.2           & 61.1          \\ \hline
Noise            & 91.1          & 59.3          & 67.4          & 81.4          & 91.4          & 8.3           & 78.0          & 87.3           & 66.8           & 62.2           & 90.0           & 59.0          \\ \hline
Brightness       & 89.7          & 63.3          & 66.9          & 83.0          & 91.3          & 12.2          & \textbf{80.2} & 85.5           & 63.6           & 83.1           & 88.8           & 63.6          \\ \hline
Contrast         & 91.1          & 72.7          & 60.7          & 86.1          & 91.3          & 12.7          & 78.6          & 88.4           & 58.4           & \textbf{85.5}  & 90.3           & 63.6          \\ \hline
Perturb          & 90.1          & 63.4          & 69.5          & 81.5          & 91.7          & 6.6           & 77.3          & 88.5           & 63.6           & 83.1           & 90.1           & 55.7          \\ \hline
Rotation         & 87.4          & 59.0          & 57.9          & 75.1          & 91.2          & 5.2           & 72.1          & 78.0           & 60.4           & 62.6           & 85.5           & 54.7          \\ \hline
Scaling          & 90.8          & 59.3          & 60.8          & 78.8          & 91.3          & 7.4           & 75.3          & 91.0           & 84.1  & 68.2           & 91.0           & 61.3          \\ \hline
Deform           & 89.7          & 61.4          & 61.5          & 81.2          & 91.6          & 7.8           & 69.2          & 86.3           & 62.4           & 31.4           & 89.2           & 51.1          \\ \hline
Top4           & 91.0          & 73.5          & 83.0          & \textbf{86.5}          & 91.6          & 45.4           & 79.4          & 90.9           & 81.9           & 80.5           & 91.2           & 74.9          \\ \hline
CycleGAN         & -             & 74.7          & 76.4          & 81.2          & -             & 18.0          & 76.2          & -              & 65.3               & 66.6               & -              & 63.5              \\ \hline
\textbf{DST (ours)} & \textbf{91.3} & \textbf{80.2} & \textbf{85.4} & \textbf{86.5} & 91.4          & \textbf{65.5} & 80.0          & \textbf{92.1}  & \textbf{84.9}           & 81.3           & \textbf{91.6}  & \textbf{80.0} \\ \hline
Supervised           & -          & 91.4 \cite{zhu2019boundary}          & 88.0 \cite{jia20183d}         & 91.9*          & -          & 94.2 \cite{xiong2019fully}           & 88.6          & -           & 92.5*           & 92.5*           & -           & 91.4          \\ \hline
\end{tabular}
\end{table}


\begin{itemize}[label=$\bullet$]
\item DST augmentation performs substantially better than any one of the tested augmentations. On average, across different tasks, DST achieves 80\% generalization Dice on unseen domains. Compare to  baseline (49.8\%) and CycleGAN (63.5\%), which achieve worse generalization performance (even though e.g. CycleGAN domain adaptation got exposure to unseen domain images). 

\item In 3D MRIs, image quality and appearance augmentation had the most impact, with larger improvements coming from \textit{sharpening}, followed by  by \textit{contrast}, \textit{brightness}, and \textit{intensity perturbation}. Spatial transforms had less impact in prostate MRI compared to heart MRI where the shape, size, and orientation of heart can be very different (see Figure~\ref{fig1}).

\item In Ultrasound, main contributions came from spatial \textit{scaling}, followed by \textit{brightness}, \textit{blurring}, and \textit{contrast} augmentations (see Figure~\ref{fig1}(c)).

\item In some datasets (such as ASC), all the individual augmentations and CycleGAN perfomred very poorly ($<13\%$ Dice), whereas DST had reasonable performance. This supports our claim that comprehensive transforms are required to cover potentially large variability of the unseen data. 

\item Individual augmentation transforms may perform slightly better on some isolated cases (e.g. brightness augmentation for WHS), but on average only DST consistently shows good generalization.  Even the combination of top 4 performing augmentations (top4) is not sufficient for robust generalization. 

\item Using only simple random crop (baseline) does not generalize well to unseen datasets (with Dice dropping as much as 40\%) , which supports importance of data augmentation in general.


\item Besides the improvements on unseen domains, DST slightly improves (2.5\%) on the source domains as well (it is valuable to not degrade the  performance on the source domain).

\item DST peformance is $\sim$10\% worse compared to fully supervised methods, as they have advantages of training and testing on the same domain and more training data. This gap can be reduced by using a larger source dataset (as shown in Section~\ref{scalingup}), in which case the DST performance is comparable to the supervised methods.

\end{itemize}

\begin{figure*}[!t]
   \vspace{-3pt}
   \begin{center}
   \begin{tabular}{c}
   \includegraphics[width=12cm]{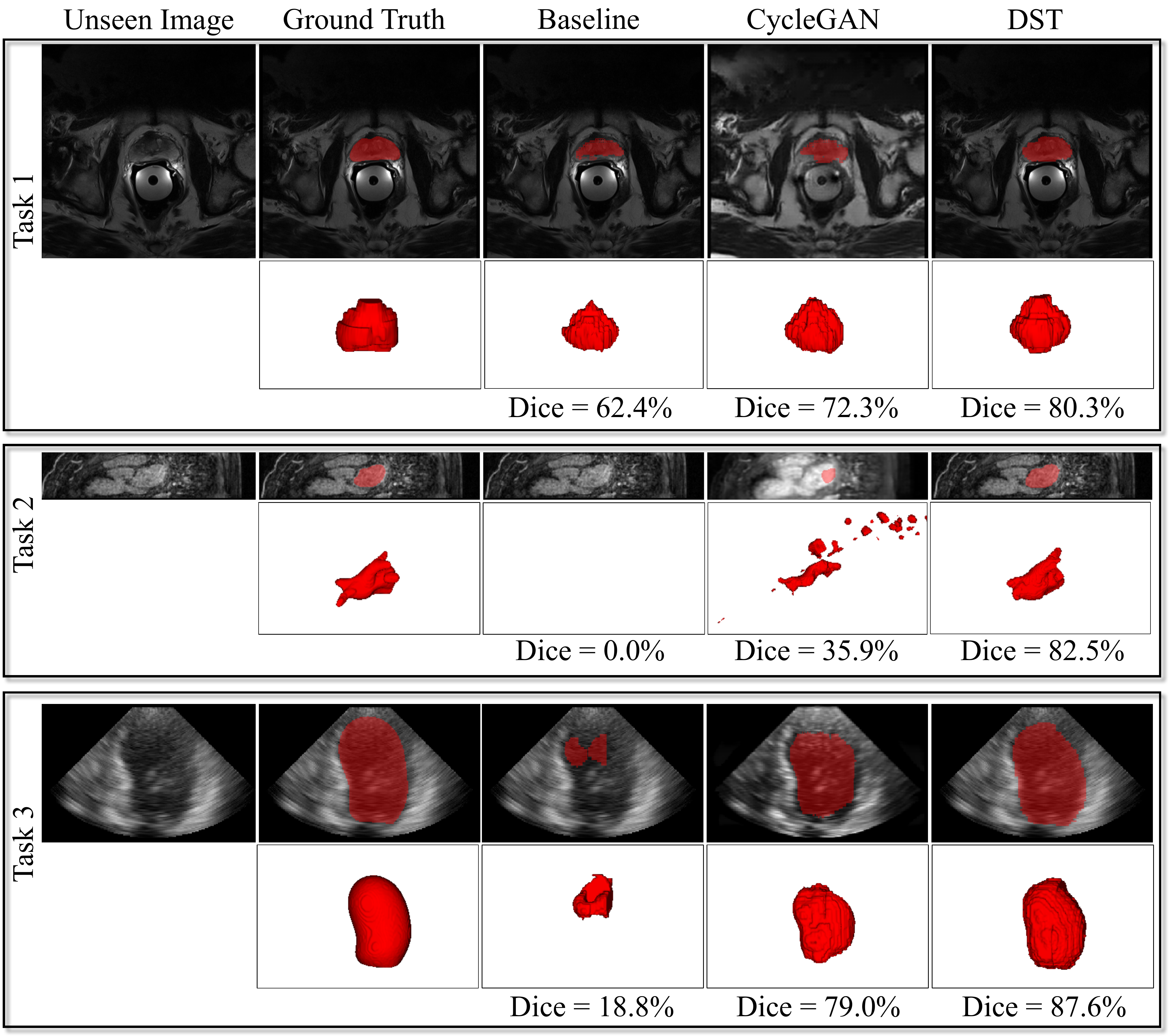}
   \end{tabular}
   \vspace{-10pt}
   \end{center}
   \caption
   { \label{figresult1} 
Generalization to unseen domains for three different 3D medical image segmentation tasks. Baseline deep models have low performances on unseen MRI and ultrasound images from different clinical centers, scanner vendors, etc. CycleGAN-based domain adaptation method helps improve segmentation performances. DST training generates robust models which significantly improve segmentation performances on unseen domains. Segmentation masks (red) overlay on unseen or CycleGAN synthesized images.
}  
   \end{figure*}

Examples of unseen domain segmentation produced by baseline model, CycleGAN-based domain adaptation, and DST domain generalization are shown in Fig. \ref{figresult1}. The baseline and DST are trained only on individual source domains, while CycleGAN requires images from target/unseen domain to train an additional generative model.


\subsubsection{DST with Larger Dataset.}
\label{scalingup}

So far we have evaluated that DST generalization performance using small ($\sim$30 volumes) public datasets. In this section, we experiment with a larger dataset, and demonstrate generalization performance comparable to supervised state-of-the-art methods.   

We train a model with DST on proprietary dataset of 465 3D MRIs (denoted as MultiCenter) with whole prostate annotations, collected from various medical centers worldwide. Table~\ref{scaleup} show the results on unseen datasets. Overall, using a large source dataset,  DST produces competitive results: with Dice being only 0.8\% lower than state-of-the-art supervised methods. 
Supervised models were trained on the same domain individually, where we were able to achieve similar performance training only on the source domain. Importantly, on the unseen domain, our DST model achieves the same performance as two radiologists (relative novice versus expert) -- it achieves a Dice score of 91.9\% on the unseen ProstateX dataset, compared with the Dice score between a novice versus expert radiologist annotations on the same dataset (also 91.9\%).
These findings suggest feasibility of practical application of deep learning models in clinical sites, where the trained DST model generalize well to unseen data.

\begin{table}[!t]
\caption{The effect of DST with larger data (465 3D MRI) for the task of whole prostate segmentation. Methods marked with * are trained and tested on the same domain or inter-observer variability (91.9\%). No evaluation of whole prostate segmentation available in MSD challenge.}
\label{scaleup}
\footnotesize
\vspace{-3pt}
\centering

\begin{tabular}{|l|c|c|c|c|c|c|c|}
\hline
                 & \multicolumn{2}{c|}{Source} & \multicolumn{5}{c|}{Unseen}                                                         \\ \hline
                 & train         & val         & MSD-P               & PROMISE       & NCI-ISBI      & ProstateX     & Average       \\ \hline
Baseline          & 95.6          & 89.9        & 87.8                & 82.9          & 88.8          & 90.6          & 87.5          \\ \hline
DST (ours)          & 94.1          & 91.8        & 89.1                & 88.1          & \textbf{89.4} & \textbf{91.9} & 89.6          \\ \hline
State-of-the-art & -             & -           & - & \textbf{91.4*} \cite{zhu2019boundary} & 88.0* \cite{jia20183d}          & \textbf{91.9*}          & \textbf{90.4} \\ \hline
\end{tabular}
\vspace{-10pt}
\end{table}

\section{Conclusion}
We propose deep stacked transformations (DST) augmentation approach for unsupervised domain generalization in 3D medical image segmentation. We evaluate DST and different augmentation strategies on three segmentation tasks (prostate 3D MRI, left atrial 3D MRI and left ventricle 3D ultrasound) when applied to unseen domains. The experiments establish a strong benchmark for the study of domain generalization in medical imaging. Furthermore, using a larger training dataset, we show that DST generalization performance is comparable to fully supervised state-of-the-art methods, making deep learning segmentation more feasible in practise. 


\bibliographystyle{splncs04}
\bibliography{ref}

\end{document}